\documentclass[10pt,twocolumn,letterpaper]{article}

\usepackage[pagenumbers]{wacv} % To force page numbers, e.g. for an arXiv version

\usepackage{graphicx}
\usepackage{amsmath}
\usepackage{amssymb}
\usepackage{booktabs}

\usepackage{mathrsfs}
\usepackage{bbding}
\usepackage{pifont}

\usepackage[pagebackref,breaklinks,colorlinks]{hyperref}

\usepackage[capitalize]{cleveref}
\crefname{section}{Sec.}{Secs.}
\Crefname{section}{Section}{Sections}
\Crefname{table}{Table}{Tables}
\crefname{table}{Tab.}{Tabs.}

\newcommand{\mysection}[1]{\vspace{2pt}\noindent\textbf{#1}}

\usepackage{adjustbox}
\usepackage{tabularx}

\def\wacvPaperID{2395} % *** Enter the WACV Paper ID here
\def\confName{WACV}
\def\confYear{2025}

\usepackage{eccvabbrv}

\usepackage{bookmark}

\begin{document}

\global\long\def\aCamera{\kappa}%

\global\long\def\focal#1{f_{#1}}%

\global\long\def\distortionFunction{\mathcal{L}}%

\global\long\def\transposed{\mathsf{T}}%

\global\long\def\transpositionSymbol{T}%

\global\long\def\translationVector{\mathbf{\mathbf{t}}}%

\global\long\def\cameraFocalPoint{\boldsymbol{C}}%

\global\long\def\cameraTripodPoint{\boldsymbol{T}}%

\global\long\def\cameraFocalPointCoordinate#1{C_{#1}}%

\global\long\def\time{t}%

\global\long\def\pan{\phi}%

\global\long\def\tilt{\theta}%

\global\long\def\roll{\gamma}%

\global\long\def\projectionMatrix{\mathtt{P}}%

\global\long\def\rotationMatrix{\mathtt{R}}%

\global\long\def\calibrationMatrix{\mathtt{K}}%

\global\long\def\identityMatrix{\mathtt{I}}%

\global\long\def\principalPoint{p}%

\global\long\def\principalPointX{\principalPoint_{x}}%

\global\long\def\principalPointY{\principalPoint_{y}}%

\global\long\def\skew{s}%

\global\long\def\distanceToPrincipalPoint{r}%

\global\long\def\coordinateImagePlaneX{x}%

\global\long\def\coordinateImagePlaneY{y}%

\global\long\def\coordinateNormalizedImagePlaneX{\overline{\coordinateImagePlaneX}}%

\global\long\def\coordinateNormalizedImagePlaneY{\overline{\coordinateImagePlaneY}}%

\global\long\def\radialDistortionParameterOne{k_{1}}%

\global\long\def\radialDistortionParameterTwo{k_{2}}%

\global\long\def\norm#1{\left\lVert #1\right\rVert _{2}}%

\global\long\def\class{c}%

\global\long\def\polyline{P}%

\global\long\def\estimatedPolyline{\hat{l}}%

\global\long\def\location{x}%

\global\long\def\pointOfEstimatedClass{\hat{\location}_{\class}}%

\global\long\def\threshold{\tau}%

\global\long\def\completenessRatio{\text{CR}}%

\global\long\def\time{t}%

\global\long\def\pitchWidth{W}%

\global\long\def\pitchLength{L}%

\global\long\def\pitchElevation{e}%

\global\long\def\tn{\mathrm{tn}}%

\global\long\def\fp{\mathrm{fp}}%

\global\long\def\fn{\mathrm{fn}}%

\global\long\def\tp{\mathrm{tp}}%

\global\long\def\numN{\mathsf{N}}%

\global\long\def\numP{\mathsf{P}}%

\global\long\def\numTN{\mathsf{T}\numN}%

\global\long\def\numFP{\mathsf{F}\numP}%

\global\long\def\numFN{\mathsf{F}\numN}%

\global\long\def\numTP{\mathsf{T}\numP}%

\global\long\def\proportion{\mathsf{p}}%

\global\long\def\proportionOfnumTN{\proportion\numTN}%

\global\long\def\proportionOfnumFP{\proportion\numFP}%

\global\long\def\proportionOfnumFN{\proportion\numFN}%

\global\long\def\proportionOfnumTP{\proportion\numTP}%

\global\long\def\proba#1{P\left(#1\right)}%

\global\long\def\clazz{y}%

\global\long\def\cpos{c^{+}}%

\global\long\def\cneg{c^{-}}%

\global\long\def\estimate#1{\widehat{#1}}%

\global\long\def\metric{\text{I}}%

\global\long\def\ppv{\text{PPV}}%

\global\long\def\npv{\text{NPV}}%

\global\long\def\tpr{\text{TPR}}%

\global\long\def\tnr{\text{TNR}}%

\global\long\def\fprE{\mbox{\ensuremath{\left(1-\tnr\right)}}}%

\global\long\def\fpr{\text{FPR}}%

\global\long\def\fnr{\mbox{\ensuremath{\left(1-\tpr\right)}}}%

\global\long\def\fprx{\text{FPR}}%

\global\long\def\fnrx{\text{FNR}}%

\global\long\def\accuracy{\text{A}}%

\global\long\def\errorrate{\text{ER}}%

\global\long\def\balancedaccuracy{\text{BA}}%

\global\long\def\jaccard{\text{JaC}}%

\global\long\def\auc{\text{AUC}}%

\global\long\def\bias{\text{B}}%

\global\long\def\priorSymbol{\pi}%

\global\long\def\priorneg{\priorSymbol{}^{-}}%

\global\long\def\priorpos{\priorSymbol^{+}}%

\global\long\def\rateOfNegativePredictions{\tau^{-}}%

\global\long\def\rateOfPositivePredictions{\tau^{+}}%

\global\long\def\Fscore{\text{F}_{1}}%

\global\long\def\recall{\text{R}}%

\global\long\def\precision{\text{P}}%

\global\long\def\cardinality{\#}%

\global\long\def\comma{\,,}%

\global\long\def\point{\,.}%

\global\long\def\wcdataset{\text{WC14}}%

% Metric

\global\long\def\iou{\text{IoU}}%

\global\long\def\ioupart{\iou_{\text{part}}}%

\global\long\def\iouwhole{\iou_{\text{whole}}}%

\global\long\def\soccerFieldTemplate{\mathcal{F}}%

\global\long\def\setOfSemanticElements{\mathcal{S}}%

\global\long\def\numberOfSemanticElements{\cardinality\allClasses}%

\global\long\def\aSemanticElement{s}%

\global\long\def\aPointOfASemanticElement{p}%

\global\long\def\aPointInThreeD{\boldsymbol{X}}%

\global\long\def\aPointInTheTwoDImage{\boldsymbol{x}}%

\global\long\def\projectionSymbol{\pi}%

\global\long\def\projectionOfAThreePoint#1#2{\projectionSymbol_{#2}\left(#1\right)}%

\global\long\def\calibrationMetric{\jaccard}%

\global\long\def\calibrationMetricWithThreshold#1{\calibrationMetric_{#1}}%

\newcommand{\ourMethodName}{BroadTrack\xspace}
\newcommand{\sncalibration}{sn-calibration\xspace}
\newcommand{\sngamestate}{sn-gamestate\xspace}

\title{BroadTrack:  Broadcast Camera Tracking for Soccer}
\author{Floriane Magera$^{1,2}$
\and
Thomas Hoyoux$^{1}$\and 
Olivier Barnich$^{1}$ \and
Marc Van Droogenbroeck$^{2}$ \and  $^1$ {\small EVS Broadcast Equipment}
 \and $^2$ {\small University of Li{\`e}ge, Belgium}
 }

\maketitle
\begin{abstract}

Camera calibration and localization, sometimes simply named camera calibration, enables many applications in the context of soccer broadcasting, for instance regarding the interpretation and analysis of the game, or the insertion of augmented reality graphics for storytelling or refereeing purposes. To contribute to such applications, the research community has typically focused on single-view calibration methods, leveraging the near-omnipresence of soccer field markings in wide-angle broadcast views, but leaving all temporal aspects, if considered at all, to general-purpose tracking or filtering techniques. Only a few contributions have been made to leverage any domain-specific knowledge for this tracking task, and, as a result, there lacks a truly performant and off-the-shelf camera tracking system tailored for soccer broadcasting, specifically for elevated tripod-mounted cameras around the stadium. In this work, we present such a system capable of addressing the task of soccer broadcast camera tracking efficiently, robustly, and accurately, outperforming by far the most precise methods of the state-of-the-art. By combining the available open-source soccer field detectors with carefully designed camera and tripod models, our tracking system, BroadTrack, halves the mean reprojection error rate and gains more than 15\% in terms of Jaccard index for camera calibration on the SoccerNet dataset. Furthermore, as the SoccerNet dataset videos are relatively short (30 seconds), we also present qualitative results on a 20-minute broadcast clip to showcase the robustness and the soundness of our system.

\end{abstract}

\section{Introduction\label{sec:tracking:Introduction}}

Sports content has the advantage of displaying sports fields, which have strictly regulated shapes and dimensions. This particularity makes sports camera calibration and localization possible in the wild, without the need for any other prior knowledge but the ability to correctly detect these field markings. 
This advantage is well-exploited, as all camera calibration techniques detect field markings, and single-frame camera calibration methods flourish, with ever-increasing accuracy. However, two facts that tend to be overlooked are (1) sports field markings can be really sparse in some areas, and (2) wide-angle broadcast cameras have typically $25\times$ zoom lenses, which can sometimes lead to quite narrow views. Thus, while the open-source datasets keep the task conveniently framed, there is an unmet need to address the case where few or even none of the field markings are visible for real-world applications.
By carefully modeling broadcast cameras,  we first include lens distortion effects ---something that is systematically overlooked in sports field registration techniques that estimate homographies---, and which we show has a great impact on the accuracy of the results. Then, by including constraints on the camera movements that a tripod actually allows, we demonstrate that with more than accurate reprojections in the image, our models recover consistent position and rotation values, leading to effectively smooth tracking. 
Besides its novel state-of-the-art performance, our tracking system is robust and includes a reinitialization strategy that is thoroughly validated on the SoccerNet-calibration dataset. Finally, our system is usable in practice, as it works from the first frame and performs at a speed of 16 frames per second (fps) for HD (1920$\times$1080 pixel resolution) images on a server equipped with two RTX 4090 GPUs, without any optimization; by experience, we know that a proper optimization of our code will make it real-time. 

To validate our technique, we use the SoccerNet datasets~\cite{Giancola2018SoccerNet,Deliege2021SoccerNetv2,Cioppa2022Scaling}.
The SoccerNet initiative has allowed new tracks of research for many
problems related to soccer video understanding, including camera calibration
and localization, by providing task definitions, corresponding public
datasets, and metrics. Among others, the SoccertNet-calibration dataset (denoted as \sncalibration hereafter) is the first open-source dataset that does not use homographies as annotations, but rather the soccer field markings and goal posts.
Besides the calibration task, which is about single-frame camera calibration, the novel game state reconstruction task~\cite{Somers2024SoccerNetGameState} aims to continuously estimate the bird's-eye view of the soccer field with the players position. The dataset of the game state reconstruction task, denoted by \sngamestate in the following, consists of soccer videos. Its annotations follow the same conventions defined for \sncalibration. Solving the game state task requires precise camera calibration and localization, consistently over time, which is a property also required for most applications regarding player performances and game analysis. 

\medskip

\mysection{Contributions.}
By leveraging specific knowledge about broadcast cameras, we produce
an efficient and accurate tracking system for soccer. We summarize
our contributions as follows: \textbf{(i)} We propose a tracking system,
named \ourMethodName for the calibration of moving cameras that
is both efficient and state-of-the-art, \textbf{(ii)} We define a
camera model that is tailored for broadcast applications, and \textbf{(iii)} We release the source code of our tracking system at  \url{https://github.com/evs-broadcast/BroadTrack}.

\section{Related work\label{sec:tracking:Related-work}}

Common approaches to solving camera calibration and localization problems
include structure from motion (SfM)~\cite{Schonberger2016Structure} and simultaneous localization and mapping (SLAM)~\cite{MurArtal2015ORBSLAM}, both relying on the camera parallax to derive properties of the 3D world in which the camera moves. A typical pipeline could be described as follows: features are detected in the video frames, matched together across time, and finally triangulated. The resulting 2D-3D correspondences are fed in a PnP solver to derive the camera parameters. Compared to the usual problems that both SfM and SLAM solve, broadcast applications require custom approaches. Indeed, real-world applications often require a metric reconstruction of the scene, but for sports applications, the world reference system should not be placed regarding a camera; rather, the cameras should all be expressed relatively to the sports field.

\paragraph{Detectors.}

Unlike common camera localization or calibration schemes, sports camera
calibration algorithms do not usually rely on 2D matching of features
between frames, but rather on the detection of sports field elements
that provide direct metric correspondences, which is an advantage
for tracking, in effect reducing the risk of drifting. Most methods
detect field markings in the image, either by binary segmentation or Hough
line transforms~\cite{Dubrofsky2008Combining,AlemanFlores2014Camera,Farin2004Robust}, and sometimes even deriving vanishing points associated to horizontal
and vertical lines~\cite{Homayounfar2017Sports,Hayet2005Fast2D}.
Lately, thanks to the performance of deep neural networks, higher
levels of semantic interpretation were attained with sports markings
being used as zone delimiters~\cite{Sha2020EndtoEnd,Tarashima2021Sports},
dividing the sports field into segmentation zones, or with some methods
identifying specific markings as unique classes. Finally, some methods derive directly sports field lines intersections because most of the methods relying on solvers for camera parameters solely use point correspondences. 

\paragraph{Virtual augmentation of correspondences.}

A global problem for all single-frame camera calibration and localization methods is the sparse nature of sports field markings, which are rarely uniformly distributed, leading to a lack of visual support in the image as there are few visible markings. Several methods alleviate this by detecting virtual correspondences on the field, or by deriving additional geometric cues from existing elements. For example, some works extend line segments to get new intersections~\cite{GutierrezPerez2024NoBells,Tsurusaki2021Sports,Cuevas2020Automatic, Falaleev2023Sportlight}, derive tangents to circles or circle keypoints~\cite{GutierrezPerez2024NoBells,Andrews2024FootyVision,Cuevas2020Automatic, Falaleev2023Sportlight}, detect grass mowing patterns~\cite{Cuevas2022Grass}, or even learn
to detect grids of keypoints on the sports field surface~\cite{Nie2021Robust,Maglo2022KaliCalib-arxiv,Maglo2023Individual,Claasen2023Video,Chu2022Sports}.
These methods leveraging grid-like keypoints on sports fields have the drawback of first relying on homographies to derive the grid projection in the image, which, as shown in previous works~\cite{Magera2024AUniversal}, are not the best fit to model the projective transformation.

\paragraph{Dictionary methods.}
Other works leverage directly prior knowledge about the broadcast
camera, such that the number of correspondences in the image does not
matter. These methods construct a dictionary of plausible broadcast
camera views, and then retrieve camera parameters based on the similarity
between the dictionary view and the camera view~\cite{Sha2020EndtoEnd,Sharma2018Automated,Chen2019Sports,Zhang2021AHigh,DAmicantonio2024Automated,Tarashima2021Sports}.
These methods tend to be slow due to the search time in the dictionary, or if the dictionary is sparsely populated, relies on the ability of a Spatial Transformer Network to regress the parameters of the homography that maps the dictionary view to the current view~\cite{Sha2020EndtoEnd,Tarashima2021Sports}.

\paragraph{Homography regression.}
Inspired by PoseNet~\cite{Kendall2015PoseNet} and learned homography
estimation~\cite{DeTone2016DeepImage-arxiv}, previous works directly
regress homographies~\cite{Fani2021Localization-arxiv,Jiang2020Optimizing,Shi2022Self}.
While the work of DeTone \etal~\cite{DeTone2016DeepImage-arxiv} estimates
the homography between a pair of images, its transposition to sports
field registration requires the regressed homography to capture the
transformation between a broadcast image and the synthetic bird's-eye view
of the sports field. The huge perspective difference between a  bird's-eye view
of the sports field and one broadcast image probably explains why these
methods rarely work in a single forward pass, limiting their use in
actual broadcast scenarios.

\paragraph{Tracking.}
Due to the limited number of datasets with broadcast videos, few methods
took interest in the temporal consistency of the camera calibration
and localization. Among these methods, we define two main categories: (a) the ones based on homographies between successive frames, and (b) the ones using traditional tracking filters. These categories are further detailed hereafter. 

\smallskip
\mysection{(a) Homography between successive frames:} 
A common strategy to perform tracking is to extract features to obtain point correspondences, or even line ones~\cite{Hayet2004Incremental}, and robustly estimate a homography between pairs of frames with RANSAC~\cite{Gupta2011Using,Maglo2023Individual}.
The underlying assumption made when computing homographies between successive frames is that the camera is purely in rotation, an assumption that other works refute for soccer~\cite{Chen2015Mimicking}. Note that a homography can be computed between frames if the correspondences are taken only on the sports field plane~\cite{Claasen2023Video,Hayet2004Incremental}.
In the case of rather small sports fields, like tennis, Farin \etal~\cite{Farin2004Robust} assume that the camera velocity is constant, and thus assume that the homography between the future image pair will be equivalent to the homography mapping the present image pair. 

\smallskip
\mysection{(b) Common temporal filters:}
A range of methods also apply common tracking algorithms such as the Kalman Filter, starting with Claasen and de Villiers ~\cite{Claasen2023Video}, who use a Kalman filter to track point correspondences on the sports field, which are later used to track the homography using an extended Kalman filter. Beetz \etal~\cite{Beetz2007Visually} use an iterative extended Kalman filter to track their camera parameters, while Citraro \etal~\cite{Citraro2020Realtime} use a particle filter to model the motion of the camera. Lu \etal~\cite{LU2019PanTiltZoom} propose a SLAM algorithm tailored for a PTZ broadcast camera by building a map of 3D rays rather than 3D points, but consider the camera focal point fixed and neglect lens distortion.

Besides NBJW~\cite{GutierrezPerez2024NoBells} which is the SOTA on the \sncalibration dataset, the methods that inspire us in terms of broadcast modeling are the ones that do model radial distortion~\cite{AlemanFlores2014Camera,Beetz2007Visually,Theiner2023TVCalib,Carr2012Pointless}
and the tripod of the camera~\cite{LU2019PanTiltZoom,Chen2018ATwopoint,Chen2015Mimicking}.

\paragraph*{Available data.}

For completeness, we addressed both camera calibration
and sports field registration techniques. However, in terms of datasets,
given the recent concerns about the ability of homographies to properly
model broadcast cameras\cite{Theiner2023TVCalib,Magera2024AUniversal},
we do not use any of the datasets that provide homographies as pseudo
ground truth~\cite{Homayounfar2017Sports,Chu2022Sports,Claasen2023Video}.
In this work, by combining high-level semantic analysis of the sports
field marking detection and broadcast camera knowledge, we derive
a new tracking system that achieves SOTA results on the \sngamestate
dataset. 

\section{Method\label{sec:tracking:Method}}

First, we define our models for the camera and for the tripod (\Cref{subsec:tracking:Method:camera}), 
then we outline our tracking system (\Cref{subsec:tracking:Method:tracking}), which comes with a reinitialization algorithm (\Cref{subsec:tracking:Initialization-algorithm}).

\subsection{Broadcast camera model} \label{subsec:tracking:Method:camera}

To model our camera, we start from the pinhole model with the \emph{calibration}
\emph{matrix} $\calibrationMatrix$, defined as follows~\cite{Hartley2004Multiple}
\begin{equation}
\mathbf{\calibrationMatrix}=\begin{bmatrix}f & 0 & \principalPointX\\
0 & f & \principalPointY\\
0 & 0 & 1
\end{bmatrix}\point
\end{equation}
As stated in the standards of telecommunications for HD~\cite{ITU2015ParametersHDTV}
and UHD content~\cite{ITU2015ParametersUHDTV}, broadcast pixels
are squares, such that the focal length is the same for both 
axes, and the skew parameter can be ignored. We further assume that
the principal point is located at the image center, so that the set of intrinsic parameters reduces to one parameter, $\focal{}$.

The camera pose is modeled by its focal point position $\cameraFocalPoint=(\cameraFocalPointCoordinate x,\cameraFocalPointCoordinate y,\cameraFocalPointCoordinate z)$
and a rotation matrix $\rotationMatrix$ parameterized by the\emph{
pan} $\pan$, \emph{tilt $\tilt$,} and \emph{roll} $\roll$ angles,
defined accordingly in the Euler angles convention: $\rotationMatrix=\rotationMatrix_{\boldsymbol{z}}(\roll)\rotationMatrix_{x}(\tilt)\rotationMatrix_{\boldsymbol{z}}(\pan)$.
In the absence of lens aberrations, the projection matrix of our camera
would be formulated by the following equation\cite{Hartley2004Multiple}:
$\projectionMatrix=\calibrationMatrix\rotationMatrix\left[\begin{array}{cc}
\identityMatrix & -\cameraFocalPoint\end{array}\right]$. However, it is not uncommon for broadcast cameras to display some
radial distortion, a deformation effect that occurs due to the curved
nature of the camera lens. While it may be modeled in the pixel space~\cite{Theiner2023TVCalib,Carr2012Pointless},
we follow the standard way and apply this deformation in normalized coordinates~\cite{Hartley2004Multiple,Zhang2000AFlexible}.

Given a point in the normalized image plane $\boldsymbol{\overline{x}}=(\coordinateNormalizedImagePlaneX,\,\coordinateNormalizedImagePlaneY)$,
the distortion function $\distortionFunction(r)=1+\radialDistortionParameterOne r^{2}$
transforms $\boldsymbol{\overline{x}}$ according to its distance
to the origin of the image plane $r=\norm{\boldsymbol{\overline{x}}}$
which finally gives the image point expressed in pixel coordinates:
\begin{equation}
\boldsymbol{x}=\focal{}\,\distortionFunction(r)\boldsymbol{\overline{x}}+\boldsymbol{\principalPoint}\point
\end{equation}
We use the model proposed by Brown-Conrady~\cite{Brown1966Decentering}, and
discard all higher orders of radial distortion and tangential distortion as we experimentally find them superfluous.
The set of unknown parameters for our camera is $\aCamera:\{\focal{},\radialDistortionParameterOne,\pan,\tilt,\roll,\cameraFocalPointCoordinate x,\cameraFocalPointCoordinate y,\cameraFocalPointCoordinate z\}$,
and we define the function $\projectionSymbol_{\aCamera}:\aPointInThreeD\rightarrow\aPointInTheTwoDImage$
that projects a 3D point to its respective image point. In the next
section, we further analyze the specificities of broadcast cameras,
and extend our model to consider the tripod on which a broadcast
camera is installed, and the constraints that it creates on the
camera parameters.

\subsubsection{Pan-tilt head and tripod\label{subsec:tracking:Method:Pan-tilt}}

Compared to the usual cameras used in the computer vision literature for
robotics or augmented reality with mobile phone captures, broadcast cameras are another kind of beast. A typical broadcast camera for soccer has a lens that can zoom up to 25 times, which consists of an arrangement of optics that can weigh up to a few kilograms~\cite{Fujifilm2024Optics}. When
we use the pinhole model to represent the arrangement of optics inside
those lenses, we approximate it with a single optic. While Chen  \etal~\cite{Chen2015Mimicking} assume that the camera focal point is at a fixed location inside the camera lens, we argue that this virtual projection center may not be located inside the camera and that it should  evolve along the optical axis as the camera zooms.
Therefore, our only assumption about the position of the focal point
regarding the physical camera is that the focal point is located
on the optical axis of the camera. We choose to rather constrain the
optical axis position regarding the tripod.

\begin{figure}
\begin{centering}
\includegraphics[scale=0.3]{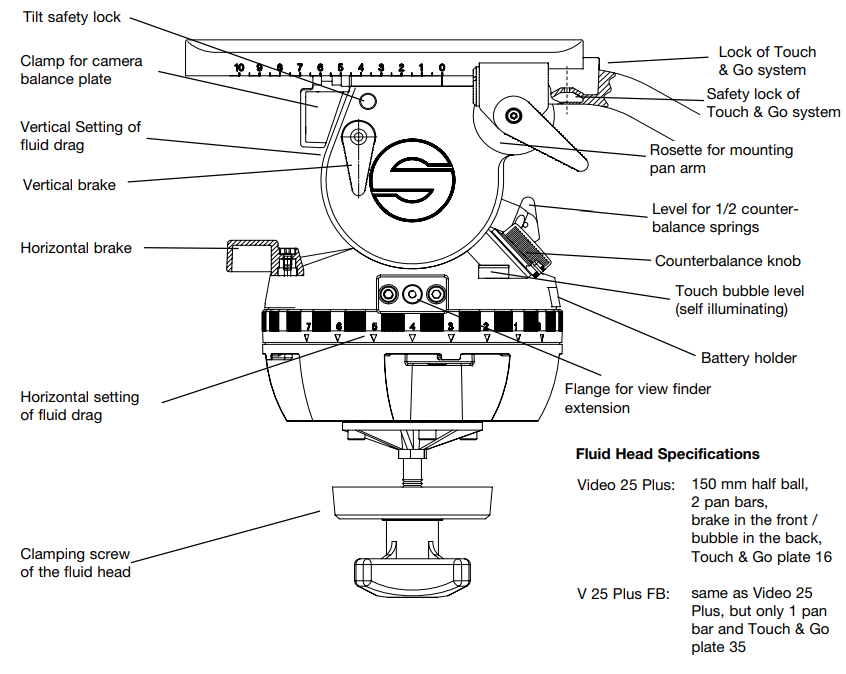}
\par\end{centering}
\caption{Usual pan-tilt head for 25$\times$ lens broadcast camera.  (taken from~\cite{Sachtler2023Video25},
others pan-tilt heads can be found online~\cite{Vinten2023Vision250}).\label{fig:tracking:Usual-pan-tilt-head}}
\end{figure}

A camera comes in several parts: the tripod, the pan-tilt head, and
the camera itself. The challenging part for our extrinsic parameters
model is the pan-tilt head. This is the part that is mounted on the
tripod on which the camera is rigged. The pan-tilt head allows
the camera to rotate smoothly while allowing the camera
operators to make quick sweeps if they lose track of the action. By
construction, it allows the camera to rotate along two axes, \ie
to pan and tilt the camera. 
To better visualize the rotation axes, a common professional pan-tilt
head is depicted in \Cref{fig:tracking:Usual-pan-tilt-head}.

\begin{figure}
\begin{centering}
\includegraphics[width=0.8\columnwidth]{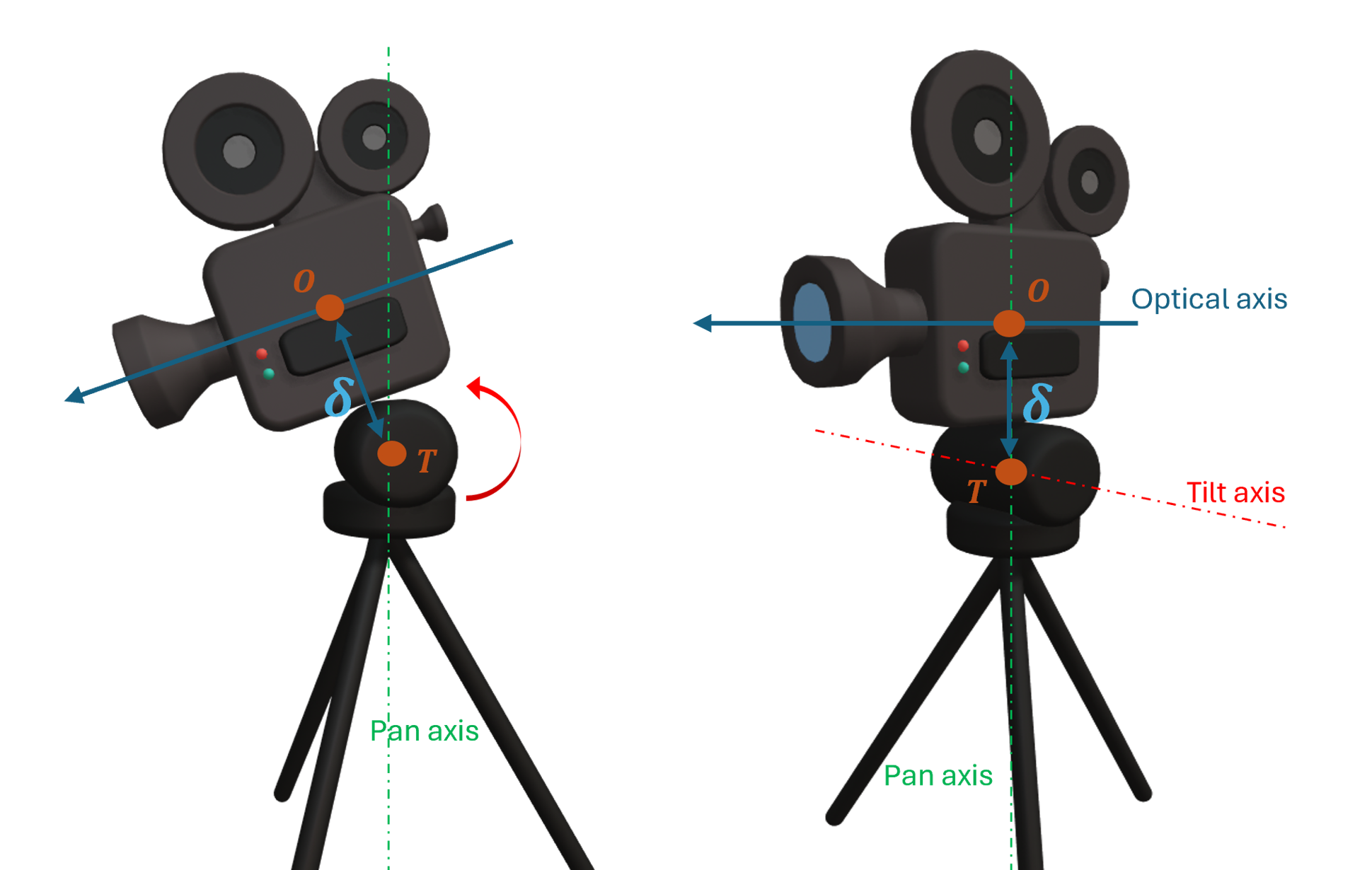}
\par\end{centering}
\caption{Tripod model. The center of rotation $\boldsymbol{T}$ remains fixed
as the camera rotates, and the point $\boldsymbol{O}$ which belongs
to the optical axis of the camera remains at a fixed $\delta$ distance
of $\boldsymbol{T}$.\label{fig:tracking:tripod-model} }
\end{figure}

According to this blueprint, we make the assumption that the pan and tilt
rotation axes intersect in a point $\cameraTripodPoint$, which remains
fixed during the whole game. Our model states that there is a
point $\boldsymbol{O}$ in the camera which belongs to the optical
axis of the camera, and which remains at a fixed distance $\delta$
of the rotation center $\cameraTripodPoint$ as the camera moves.
Let $\boldsymbol{r_{i}}$ be the $i$th column of the orientation
matrix of the camera $\rotationMatrix^{\transposed}$, the vector $-\boldsymbol{r_{2}}$
is the upvector of the camera, while the vector $\boldsymbol{r_{3}}$ defines
its optical axis direction. If we further assume that the camera is
centered on the pan-tilt head, this point $\boldsymbol{O}$ is then
determined by the upvector $-\boldsymbol{r_{2}}$ and its distance
to the tripod rotation center $\boldsymbol{O}=\cameraTripodPoint-\delta\boldsymbol{r_{2}}$.
As the focal point of the camera changes with the zoom level of the
camera, the position of the focal point is finally given by $\cameraFocalPoint=\cameraTripodPoint-\delta\boldsymbol{r_{2}}+\lambda\boldsymbol{r_{3}}$.
The parameters $\left\{ \cameraTripodPoint,\delta\right\} $ are fixed
over time, while $\lambda$ varies with the camera view. A visualization
of the model is shown in \Cref{fig:tracking:tripod-model}.

\subsection{Complete description of our tracking system}
\label{subsec:tracking:Method:tracking}

Our system comprises different steps, described hereafter: the
detection of the field markings (\Cref{subsec:Sports-Field-Detection}),
which is augmented using optical flow (\Cref{subsec:Optical-Flow}), a parameter update
procedure (\Cref{subsec:tracking:Method:Optimization}), and an
evaluation of the tracking confidence (\Cref{subsec:Online-confidence-evaluation}).

\subsubsection{Sports field detection\label{subsec:Sports-Field-Detection}}

Let $\setOfSemanticElements$ be the set of semantic classes constituent
of soccer field markings, such as ``left goal line'', ``center circle'', \etc
Each one can be viewed as a simple geometric element. The laws of
the game~\cite{IFAB2022Laws} further specify their position and
dimensions, such that we can define the soccer field template $\soccerFieldTemplate:\setOfSemanticElements\rightarrow\mathtt{E}$,
which maps each semantic class $c \in \setOfSemanticElements$ to its corresponding 3D geometric element $e \in \mathtt{E}$.
For convenience, the projection of the soccer field markings in the
camera view $\aCamera$ will be denoted by $\projectionSymbol_{\aCamera}(\soccerFieldTemplate)$.

To obtain 2D-3D correspondences, we only need to detect and
identify soccer field elements in the videos. We leverage existing
open-source sports field detectors such as the keypoint detector of
Gutiérrez-Pérez and Agudo~\cite{GutierrezPerez2024NoBells} that detects
the intersection points of the soccer field markings. 
While our reinitialization algorithm uses
the detected points, the tracking leverages the denser information retrieved from semantic segmentation of field markings. We rely on the semantic field markings
detector of Theiner and Ewerth~\cite{Theiner2023TVCalib}. Semantic
segmentation of field markings provides blobs of pixels towards 3D line or circle equation correspondences. As segmentation blobs provide a robust, but maybe too rich source of information, we synthesize it with the mean shift
algorithm~\cite{Comaniciu2002Meanshift} to fit a set of points that
approximate the segmented blob, thus generating a set of image points
$\aPointInTheTwoDImage_{c}^{1},...\aPointInTheTwoDImage_{c}^{n}$
per visible soccer field marking class $c$ of $\setOfSemanticElements$.

\subsubsection{Optical flow\label{subsec:Optical-Flow}}

As broadcast cameras can zoom in on areas of the soccer field that are sparse in terms of markings, optical flow correspondences are necessary to prevent drifting. 
Furthermore, these correspondences can be sampled uniformly in the image, which helps to distribute the visual support in the whole image, unlike field markings which can be condensed into small areas. 
We use the pyramidal version of Lucas-Kanade's algorithm~\cite{Bouguet2001Pyramidal}
to retrieve $N_{o}$ point matches $\left\{ \aPointInTheTwoDImage_{t-1}^{i}\leftrightarrow\aPointInTheTwoDImage_{t}^{i}\right\} _{i=0}^{N_{o}}$
between the previous image $\mathcal{I}_{t-1}$ and the current one $\mathcal{I}_{t}$. 

\subsubsection{Update through non-linear optimization\label{subsec:tracking:Method:Optimization}}

Starting from the previous camera estimate $\aCamera_{t-1}$, we update the camera parameters $\aCamera_t:\{\focal{},\radialDistortionParameterOne,\pan,\tilt,\roll,\cameraFocalPointCoordinate x,\cameraFocalPointCoordinate y,\cameraFocalPointCoordinate z\}$ by minimizing the sum of three error functions.

First, given soccer field markings correspondences, if we denote by $\setOfSemanticElements^{*}$
the subset of $\setOfSemanticElements$ that is detected in the current
frame $\mathcal{I}_{t}$, and by $\left\{ \aPointInTheTwoDImage_{c}^{i}\right\} _{i=0}^{N_{c}}$ the
set of points extracted by mean shift from the segmentation maps
for the soccer field marking class $c$, we minimize the reprojection
error between each extracted
point and the closest point $\textit{\textbf{p}}$ belonging to the projected soccer field element $\projectionSymbol_{\aCamera}(\soccerFieldTemplate(c))$ : 
\begin{equation}
\mathcal{L}_{\soccerFieldTemplate}=\sum_{c\in\setOfSemanticElements^{*}}\sum_{i=0}^{N_{c}}\rho\left(\min_{\textbf{\textit{p}} \in \projectionSymbol_{\aCamera}(\soccerFieldTemplate(c)) } \left\Vert \textbf{\textit{p}}-\aPointInTheTwoDImage_{c}^{i} \right\Vert_{2}\right)\comma\label{eq:tracking:reprojection-loss}
\end{equation}
 where $\rho(.)$ is the Cauchy loss, used to filter outliers. 

Secondly, given optical flow correspondences, we can filter out the point correspondences landing
outside the polygon obtained by the projection of the soccer field
side lines. To minimize the noise coming from the players
motion, the players are detected with RTMDet~\cite{Lyu2022RTMDet-arxiv}, and the correspondences are discarded around their bounding boxes. A point correspondence
$\aPointInTheTwoDImage_{t-1}^{i}$ in frame $\mathcal{I}_{t-1}$ can
be mapped to a ray by inverting the projection function: $\mathtt{L}_{t-1}^{i}=\projectionSymbol_{\aCamera_{t-1}}^{-1}(\aPointInTheTwoDImage_{t-1}^{i})\point$
As we only keep correspondences on the surface of the sports field,
its intersection with the plane $Z=0$ allows us to retrieve a 3D point
$\aPointInThreeD_{t-1}^{i}=$ $\mathtt{L}_{t-1}^{i}\left(\begin{array}{cccc}
0 & 0 & 1 & 0\end{array}\right)^{\transposed}$. As a result, for all detected optical flow correspondences, we minimize the Cauchy loss of 
the reprojection error:
\begin{equation}
\mathcal{L}_{OF}=\sum_{i=0}^{N_{o}}\rho\left( \left\Vert \projectionSymbol_{\aCamera}(\aPointInThreeD_{t-1}^{i})-\aPointInTheTwoDImage_{t}^{i}\right\Vert _{2}\right)\point\label{eq:tracking:optical-flow-loss}
\end{equation}

The third term of our objective function is a constraint that ensures
that the camera rotation and focal point position satisfy our tripod
model. We derive the position of the point $\boldsymbol{O^{*}}$ as
the closest point to $\boldsymbol{T}$ belonging to the camera optical
axis: 
\begin{equation}
\boldsymbol{O^{*}}=\cameraFocalPoint+\frac{\left\langle (\cameraTripodPoint-\cameraFocalPoint),\boldsymbol{r_{3}}\right\rangle \boldsymbol{r_{3}}}{\left\Vert \boldsymbol{r_{3}}\right\Vert _{2}^{2}}\comma\label{eq:tracking:derive-O}
\end{equation}
where $\left\langle .\comma.\right\rangle $ denotes the dot product. Since the
distance between this point and the tripod rotation center should
be $\delta$ meters, we define the following loss term:
\begin{equation}
\mathcal{L}_{T}=\delta-\left\Vert \boldsymbol{O^{*}}-\cameraTripodPoint\right\Vert _{2}\point\label{eq:tracking:tripod-constraint}
\end{equation}

The previous losses are scaled and summed into a final objective function
$\mathcal{\mathcal{L}}=\mathcal{L}_{\soccerFieldTemplate}+\mathcal{L}_{OF}+\omega\mathcal{L}_{T}$, 
with $\omega$ being a scalar hyperparameter, to optimize the camera
parameters $\aCamera$ using the Levenberg-Marquardt algorithm.

\subsubsection{Online confidence evaluation\label{subsec:Online-confidence-evaluation}}

Given the semantic segmentation of the soccer field markings $S$, we produce
a binary map $B$ of all the soccer field markings in the image: $B(x,y)=1,S(x,y)=c,\forall c\in\setOfSemanticElements$.
Considering the projection of the soccer field template $\soccerFieldTemplate$
in the image $\projectionOfAThreePoint{\soccerFieldTemplate}{\aCamera}$,
our confidence score is the Jaccard index between the two generated
masks: 
\begin{equation}
s=\dfrac{\left|B\cap\projectionOfAThreePoint{\soccerFieldTemplate}{\aCamera}\right|}{\left|B\cup\projectionOfAThreePoint{\soccerFieldTemplate}{\aCamera}\right|}\point\label{eq:tracking:confidence-score}
\end{equation}

This score is used to detect when a reinitialization of the
tracker is needed, and when the optical flow correspondences are to
be discarded to prevent them from damaging the optimization, as it relies on the previous camera parameters $\aCamera_{t-1}$.

\subsection{Reinitialization algorithm\label{subsec:tracking:Initialization-algorithm}}

For initialization, or when the confidence score becomes low, we propose a reinitialization algorithm. With the capabilities of deep neural networks today, soccer field markings detection is not a very difficult task. Usually,
the tracking starts drifting when the sports field markings become
very sparse in the image, which is why we include a reinitialization algorithm that only needs two point correspondences.

Given two point correspondences between the image and the top-view
sports field model, our reinitialization algorithm estimates the
focal length, the pan, and the tilt of the camera. The focal point is set to the tripod rotation center, and the roll value is set to
zero.

Then, given the image points
$\boldsymbol{x_{i}}=(x_{i},y_{i})$ and their corresponding world
points $\boldsymbol{X_{i}}=(X_{i},Y_{i},Z_{i})$, we derive the camera
parameters $\{\focal{},\pan,\tilt\}.$ The focal length $\focal{}$ is estimated with the equations originally 
derived by Gedikli~\cite{Gedikli2008Continual}, and summarized in Appendix 6.1 of \cite{Chen2018ATwopoint}. The pan and tilt angles, respectively $\pan$ and $\tilt$, are estimated by an iterative algorithm assuming
that $\pan$ only affects the projection along the $x$-axis, and that $\tilt$ only affects the projection along the $y$-axis. We initialize $\pan$ and $\tilt$ by steering
the optical axis towards the barycenter of the detected world
points $\boldsymbol{X}_{i}$. Then we estimate updates depending on
the correspondences:
\begin{gather}
d\pan=\sum_{i}\frac{1}{2}\tan^{-1}(x_{i}-\projectionOfAThreePoint{\aPointInThreeD_{i}}{\aCamera}_{x},\focal{})\comma
\\
d\tilt=\sum_{i}\frac{1}{2}\tan^{-1}(y_{i}-\projectionOfAThreePoint{\aPointInThreeD_{i}}{\aCamera}_{y},\focal{})\comma
\end{gather}
with $\projectionOfAThreePoint .{\aCamera}_{x}$ denoting the $x$ coordinate
of the point projection. The pan and tilt are thus iteratively refined as
$\pan_{k+1}=\pan_{k}+d\pan_{k}$ and $\tilt_{k+1}=\tilt_{k}+d\tilt_{k}$,
with $k$ being less or equal to 5.

If there are more than two correspondences in the image, we use RANSAC
to filter out potential outliers. As a last step, the optimization described in \Cref{subsec:tracking:Method:Optimization}
is performed without the $\mathcal{L}_{OF}$ error function.

\section{Results\label{sec:tracking:Results}}

To validate our tracking system, we conducted several experiments. In \Cref{subsec:tracking:results:sn-gamestate-evaluation}, we first report the performance of \ourMethodName on \sngamestate according to the metrics proposed with the \sncalibration dataset.
Then, in \Cref{tracking:results:reinit}, we validate our reinitialization algorithm on the 2023 version of the \sncalibration challenge. To further validate the effects of optical flow, tripod constraint, and lens distortion, we provide an ablation study in \Cref{subsec:tracking:Ablation-study}. Finally, since the metrics are meant for single-frame camera calibration and only evaluate the quality of the reprojection in the images, in \Cref{subsec:tracking:results:physical-interpretation}, we qualitatively illustrate  the suitability of the camera parameters derived by \ourMethodName.  

\subsection{Tracking system on SoccerNet-gamestate \label{subsec:tracking:results:sn-gamestate-evaluation}}
We evaluate our system on the \sngamestate dataset, which consists
of sequences of 30 seconds, where each frame is annotated with point
correspondences along the soccer field markings and goal posts. The
test set contains 49 sequences of 750 images each, extracted from
3 games of the Swiss Football League. 

For our experiments, we use the $\calibrationMetric$ metric~\cite{Magera2024AUniversal} (previously denoted by $\text{AC}$ in the \sncalibration challenges~\cite{Giancola2022SoccerNet,Cioppa2024SoccerNet2023Challenge}), which expresses the percentage of soccer field elements that are
correctly reprojected in the image, the correctness being tuned by a tolerance parameter $\tau$ in pixels. We report
this metric for both $5$ and $10$ pixels, which are quite challenging
threshold values for the HD frames of the dataset. For comparison,
note that the SoccerNet camera calibration challenge evaluates this
metric for 5 pixels, but for $960 \times 540$-sized images, a quarter of this dataset resolution.
We also give the mean reprojection error and the completeness rate, the latter indicating the proportion of the dataset for which the technique produced camera parameters.

To estimate the tripod rotation center $\cameraTripodPoint$,
we run our tracking system without the tripod constraint, and then optimize
the tripod position and distance to the optical axis $\delta$ given
the estimated camera parameters. From the keypoints detected by the neural network of Gutiérrez-Pérez and Agudo~\cite{GutierrezPerez2024NoBells}, we only keep the ones that are actual marks, line intersections,  or line and circle (arc) intersections, \ie we discard all keypoints that are derived from other geometric cues. The reinitialization algorithm is used to get the parameters of the first frame, and during the tracking, the reinitialization is performed when the score confidence score $s$ falls under $0.5$, which leads to frequent reinitialization.

During our evaluation, out of the 49 sequences of 750 frames in the test set, the reinitialization had to be used 33 times, and the loss of tracking lasted for 15 frames on average. From the result reported in Table~\ref{tab:tracking:results:sn-gamestate},
\ourMethodName outperforms all available open-source methods. We also
experimentally confirm that the modeling of a broadcast camera as a PTZ camera with a fixed focal point deters the performance of the tracking. This emphasizes the specificity of broadcast cameras, which are not well modeled by PTZ cameras, even if they are rigged on a tripod.

\begin{table}
\begin{centering}
\begin{adjustbox}{max width=\columnwidth}
\begin{tabular}{c|c|c|c|c}
 & $\calibrationMetricWithThreshold 5$($\uparrow$)& $\calibrationMetricWithThreshold{10}$($\uparrow$) & MRE($\downarrow$) & CR($\uparrow$)\tabularnewline
\hline 
\hline 
TVCalib\cite{Theiner2023TVCalib} & 19.88 & 50.42 & 12.4 & 99.93\tabularnewline % note : pas 100% car je filtre les résultats qui ne projettent rien dans l'image
\hline 
NBJW\cite{GutierrezPerez2024NoBells} & 37.14 & 68.24 & 10.28 & 93.67\tabularnewline
\hline 
PTZ-SLAM\cite{LU2019PanTiltZoom} & 25.87 & 45.28 & 27.64 & 26.67{*}\tabularnewline
\hline 
Ours (fixed $\cameraFocalPoint$) & 50.97 & 74.93 & 5.39 & 100\tabularnewline
\hline 
\ourMethodName & \textbf{56.88} & \textbf{79.79} & \textbf{5.02} & \textbf{100}\tabularnewline
\end{tabular}
\end{adjustbox}
\par\end{centering}
\caption{Comparison metrics on the \sngamestate dataset. MRE stands for Mean
Reprojection Error and is measured in pixels; CR stands for completeness rate in percent. All results are reported for HD, $1920\times1080$ frames.
For the PTZ-SLAM method, the code provided by the authors crashes after processing about 200 frames; hence the low completeness rate (see {*}). \label{tab:tracking:results:sn-gamestate}}
\end{table}

\subsection{Reinitialization on SoccerNet-calibration \label{tracking:results:reinit}}

We evaluate our reinitialization algorithm on the \sncalibration
dataset. The test set comprises $3{,}141$ images from a wide range of
cameras used during soccer broadcast, including wide-angle cameras, and fish-eye cameras. 
The $\calibrationMetric$ evaluation  of our method is computed only for views that obtain a confidence score $s < 0.2$, which lowers our completeness rate. 
We also expect this score condition to filter out views for which our tracking system is not especially designed, \eg views from fish-eye cameras. 
The default focal point localizations are defined for
common wide-angle cameras, that is for main, 16 meters, and high behind the
goal (\emph{HBG}) cameras respectively as $\cameraFocalPoint_{main}=(0,55,-12)$, $\cameraFocalPoint_{16m}=(\pm36,55,-12)$,
$\cameraFocalPoint_{HBG}=(-65,0,-15)$, all expressed in meters in the \sncalibration world reference system~\cite{Magera2022SoccerNet}. 

From the results reported in \Cref{tab:tracking:Two-points-reinitialization}, we establish that our algorithm can reach state-of-the-art performance on a smaller part of the dataset. Since the camera diversity of the dataset is higher than the one we design our system for, the high performance of the reinitialization part comforts our strategy of frequent reinitialization. 

\begin{table}
\begin{centering}
\begin{tabular}{c|c|c|c}

 & $\calibrationMetricWithThreshold 5$($\uparrow$) & $\calibrationMetricWithThreshold{10}$($\uparrow$) & CR($\uparrow$)\tabularnewline
\hline 
\hline 
TVCalib \cite{Theiner2023TVCalib} & 52.9 & 73.4 & 66.5
\tabularnewline
\hline
NBJW \cite{GutierrezPerez2024NoBells} & 73.7 & 86.7 & \textbf{77.5}\tabularnewline

\hline 
Ours & \textbf{75.25} & \textbf{86.8} & 69.8\tabularnewline
 
\end{tabular}
\par\end{centering}
\caption{Two-point reinitialization algorithm evaluation on the \sncalibration
dataset of 2023.  $\protect\calibrationMetric$ metrics are reported  in percent at $5$ and $10$ pixels for $960\times540$ frames. Our reinitialization algorithm equals or achieves the SOTA performance of the NBJW method on a slightly smaller subset of the dataset. To enable fairer comparison with TVCalib, we filter out their calibrations with $\protect\calibrationMetric_5 = 0$, because, by design, their method never signals failure. \label{tab:tracking:Two-points-reinitialization} }

\end{table}

\subsection{Ablation study\label{subsec:tracking:Ablation-study}}

We conduct our ablation study on the \sngamestate test set. To run our algorithm without any prior knowledge of the tripod rotation center,
we choose a default position as input to our initialization algorithm.
We arbitrarily set the camera focal point to
a default main location in $\cameraFocalPoint=(0,55,-12)$ meters.
As shown in \Cref{tab:tracking:Ablation-study}, the biggest improvement in performance comes from the inclusion of radial distortion in the
camera model, which legitimates our concerns about the previous datasets based on homographies and confirms our choice of not using their annotations as it would lead to an unfair evaluation. 
It is also worth noticing that the optimization procedure starting from the previous camera estimate does not lead to much of a performance boost when compared with the NBJW method. 
This demonstrates that their strategy of deriving virtual keypoints pays off, even if it is at the expense of physical modeling of the camera parameters, as discussed in \Cref{subsec:tracking:results:physical-interpretation}.
Both optical flow and tripod constraints demonstrate a smaller contribution to the performance of our algorithm according to the reported metrics of \Cref{tab:tracking:Ablation-study}. We argue and show in the next section that their contribution is only partially reflected by the single-view metrics used until now. In the next section, we show their benefits through qualitative evaluation and comparisons.   

\begin{table}
\begin{centering}
\begin{adjustbox}{max width=\columnwidth}
\begin{tabular}{c|c|c||c|c|c|c}
OF & $\cameraTripodPoint$ & $k_{1}$ & $\calibrationMetricWithThreshold 5$($\uparrow$) & $\calibrationMetricWithThreshold{10}$($\uparrow$) & MRE($\downarrow$) & MedRE($\downarrow$)  \tabularnewline
\hline 
\hline
\ding{56} & \ding{56} & \ding{56} & 42.09 & 74.09 & 5.74 & 3.13  \tabularnewline
\hline 
\ding{56} & \ding{56} & \CheckmarkBold{} & 54.1 & 78.35 & 5.04 & 2.47  \tabularnewline
\hline 
\CheckmarkBold{} & \ding{56} & \CheckmarkBold{} & 55.96 & 79.07 & \textbf{4.85} & 2.43  \tabularnewline
\hline 
\ding{56} & \CheckmarkBold{} & \CheckmarkBold{} & 54.9 & 78.99 & 4.95 & 2.44  \tabularnewline
\hline 
\CheckmarkBold{} & \CheckmarkBold{} & \CheckmarkBold{} & \textbf{56.88} & \textbf{79.79} & 5.02 & \textbf{2.37}  \tabularnewline
\end{tabular}
\end{adjustbox}
\caption{Ablation study on \sngamestate. $\protect\calibrationMetric$ metrics are reported  in percent at $5$ and $10$ pixels for HD frames. Mean (MRE) and Median (MedRE) Reprojection Errors are reported in pixels. OF stands for optical flow, $\textit{\textbf{T}}$ is for our tripod constraint, and finally $k_1$ represents the inclusion of radial distortion in our model.\label{tab:tracking:Ablation-study}}
\par\end{centering}
\end{table}

\subsection{Physical soundness \label{subsec:tracking:results:physical-interpretation}}

\begin{figure*}[htbp]
    \centering
    \setlength{\tabcolsep}{0pt}
    \begin{tabular}{ccc}
          \begin{subfigure}[b]{0.32\textwidth}
            \centering
            \includegraphics[width=\textwidth]{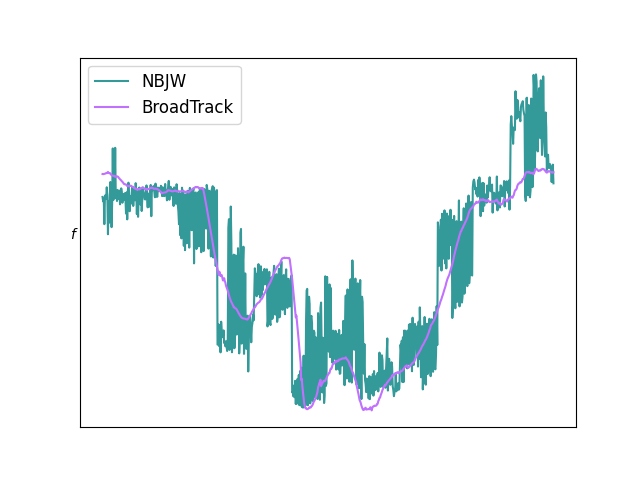}
            \caption{Focal length value variations.}
            \label{subfig:tracking:nbjw-f}
        \end{subfigure} &
        \begin{subfigure}[b]{0.32\textwidth}
            \centering
            \includegraphics[width=\textwidth]{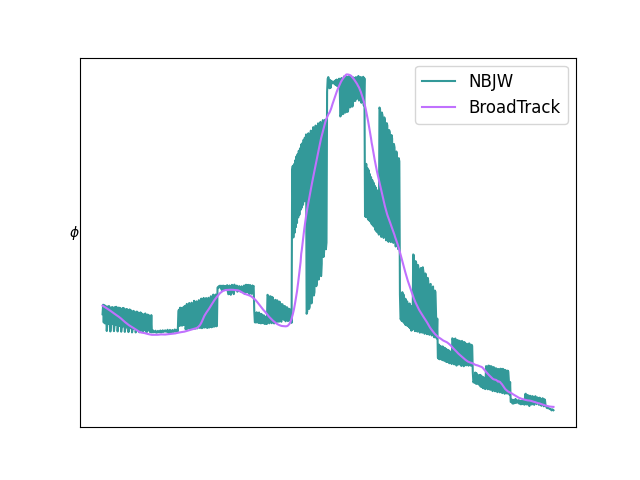}
            \caption{Pan value variations.}
            \label{subfig:tracking:nbjw-pan}
        \end{subfigure}&
        \begin{subfigure}[b]{0.32\textwidth}
            \centering
            \includegraphics[width=\textwidth]{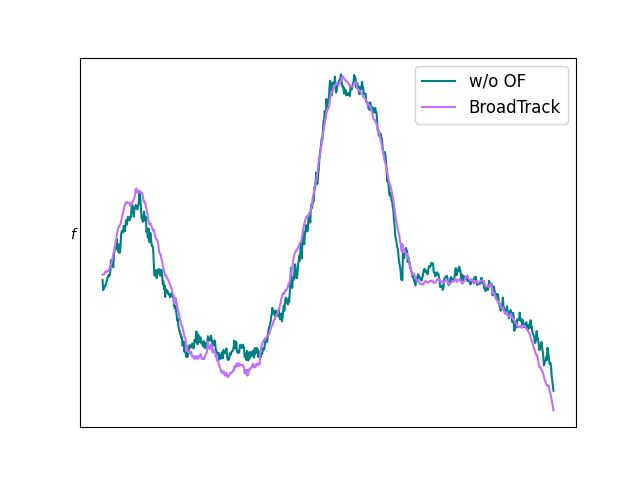}
            \caption{Focal length with or without optical flow.}
            \label{subfig:tracking:no-of-f}
        \end{subfigure} 
        % &
        
        \\
        \begin{subfigure}[t]{0.32\textwidth}
            \centering
            \includegraphics[width=\textwidth]{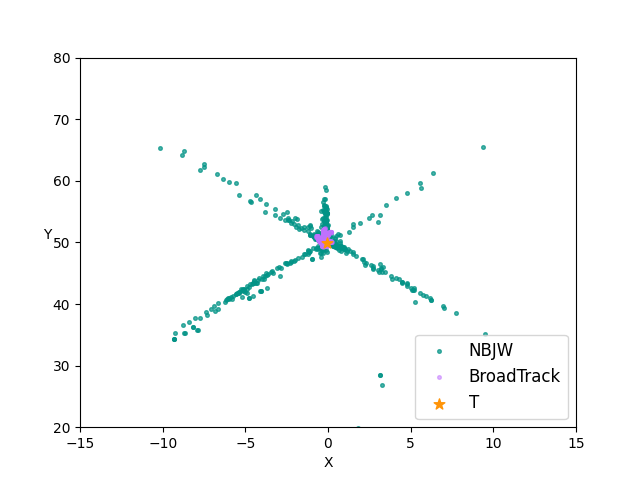}
            \caption{Focal point position in the $XY$ plane.}
            \label{subfig:tracking:nbjw-pos-1}
        \end{subfigure} &
         \begin{subfigure}[t]{0.32\textwidth}
            \centering
            \includegraphics[width=\textwidth]{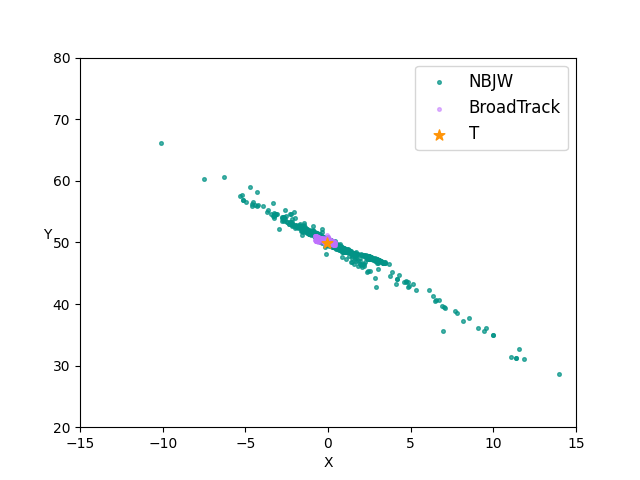}
            \caption{Focal point position in the $XY$ plane.}
            \label{subfig:tracking:nbjw-pos-2}

        \end{subfigure}&
        \begin{subfigure}[t]{0.32\textwidth}
            \centering
            \includegraphics[width=\textwidth]{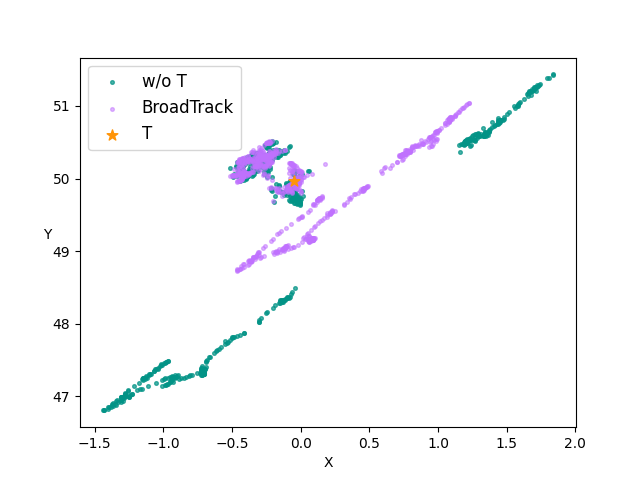}
            \caption{Benefit of including the tripod constraint on the focal point position in the $XY$ plane.}
            \label{subfig:tracking:no-T-pos-2}
        \end{subfigure} 
    \end{tabular}
    \caption{Camera parameters visualizations, best viewed on screen. The first row displays pan and focal length values along test sequences of the \sngamestate dataset. \Cref{subfig:tracking:nbjw-f} and \ref{subfig:tracking:nbjw-pan} show the jitter of the parameters extracted by NBJW compared to \ourMethodName. \Cref{subfig:tracking:no-of-f} shows that the optical flow helps to smooth focal length values. The second row displays the focal point $\cameraFocalPoint$ position in the $XY$ plane. \Cref{subfig:tracking:nbjw-pos-1} and \ref{subfig:tracking:nbjw-pos-2} show that the focal point estimated by NBJW can travel up to $20$ meters along a single sequence, while our focal point remains in a close neighborhood of the estimated tripod position. Finally, \Cref{subfig:tracking:no-T-pos-2} shows the benefit of including the tripod constraint on the camera position, which remains closer to the estimated center of rotation $\cameraTripodPoint$. 
    } 
    \label{fig:tracking:physical-parameters}
\end{figure*}

As explained in \Cref{subsec:tracking:Method:Pan-tilt}, professional pan-tilt heads are made to ensure the smooth motion of the camera. This means that we expect pan and tilt values to be particularly smooth over time. We also expect our tripod constraint to make the camera focal points more localized, even if, per se, there is no restriction in terms of distance to the tripod. These intuitions are confirmed by our visualization of \Cref{fig:tracking:physical-parameters}. We notice that, while the focal lengths estimated by \ourMethodName are smoother or display lower variations than the other methods, it still shows some high-frequency variations. We explain that because of a vertigo effect, as the scene is far away, some uncertainty in terms of position can be compensated by a focal length adjustment, and conversely, without displaying perspective distortions.  

\subsection{Long-term tracking}
\begin{figure*}[htbp]
    \centering
    \noindent\begin{tabular}{ccc}
          \begin{subfigure}[b]{0.32\textwidth}
            \centering
            \includegraphics[width=\textwidth]{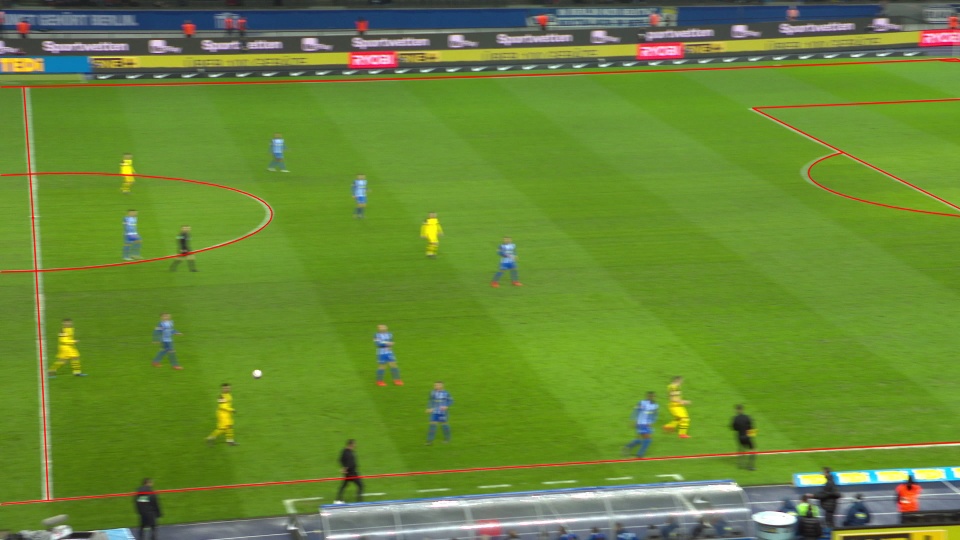}
        \end{subfigure} &
        \begin{subfigure}[b]{0.32\textwidth}
            \centering
            \includegraphics[width=\textwidth]{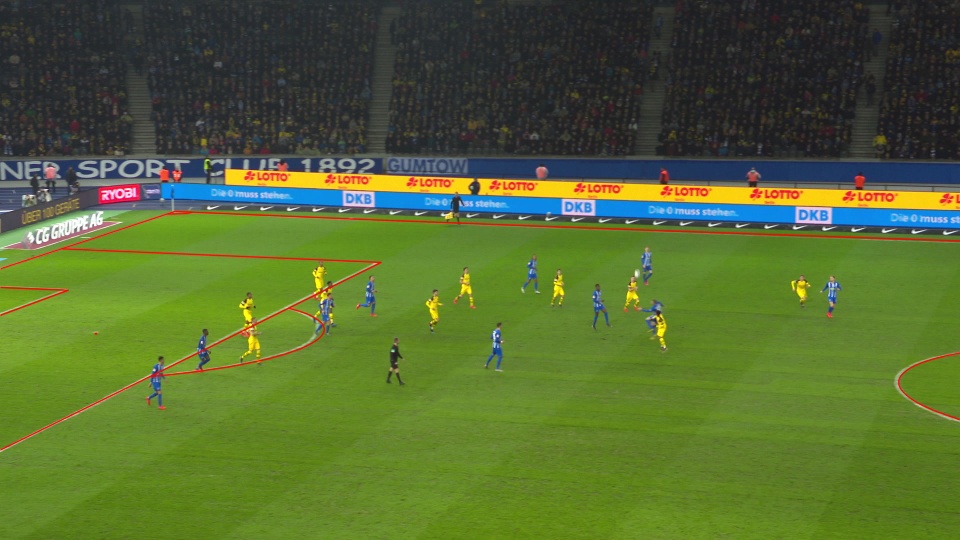}
        \end{subfigure} &
        \begin{subfigure}[b]{0.32\textwidth}
            \centering
            \includegraphics[width=\textwidth]{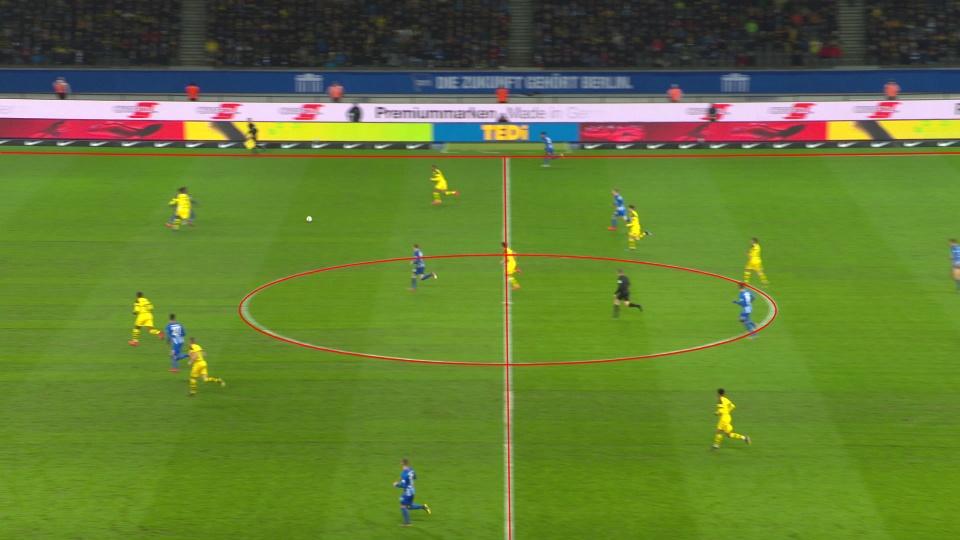}
        \end{subfigure} 
        \\
        \begin{subfigure}[b]{0.32\textwidth}
            \centering
            \includegraphics[width=\textwidth]{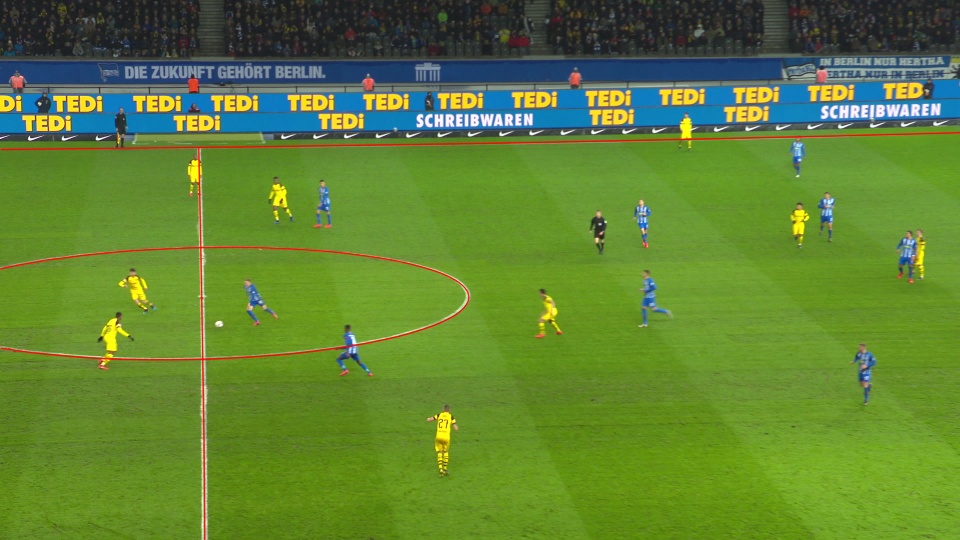}
        \end{subfigure} &
         %\\
        \begin{subfigure}[b]{0.32\textwidth}
           \centering
           \includegraphics[width=\textwidth]{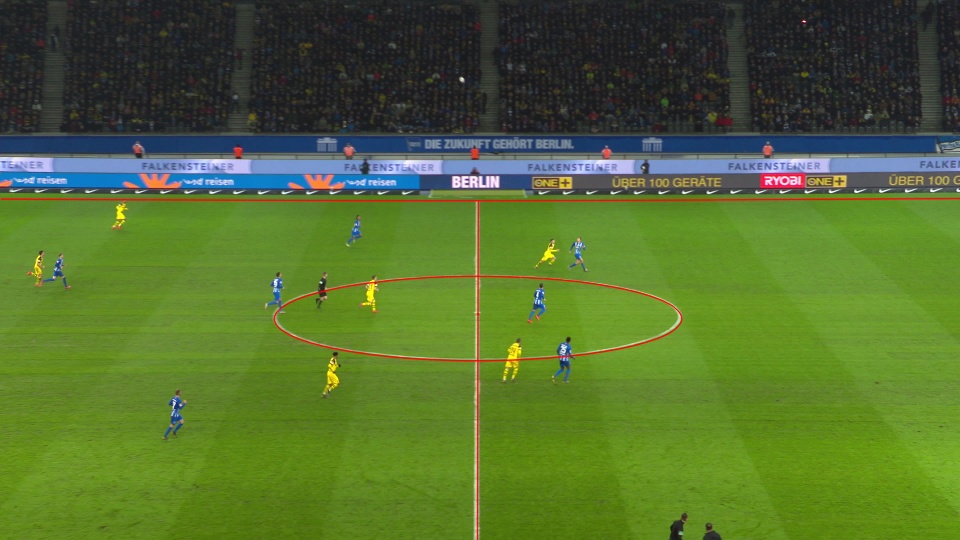}
        \end{subfigure} &
        \begin{subfigure}[b]{0.32\textwidth}
           \centering
           \includegraphics[width=\textwidth]{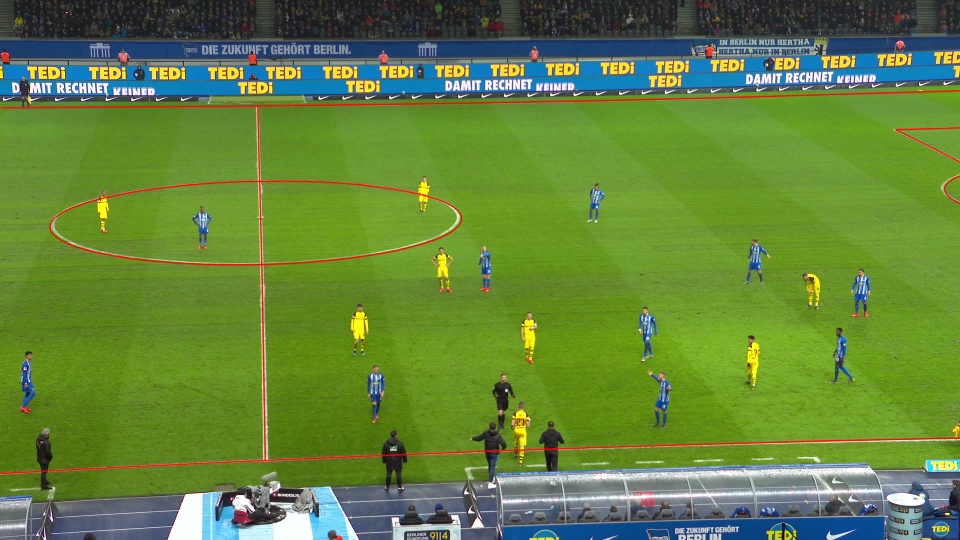}
        \end{subfigure} 
    \end{tabular}
    \caption{Qualitative evaluation of the 20 minutes sequence from the Bundesliga. One out of $10{,}000$ images is displayed. Soccer field markings are reprojected in red using the estimated camera parameters $\aCamera$. }
    \label{fig:tracking:long-sequence}
\end{figure*}

To further demonstrate the performance of \ourMethodName, we show results on 20 minutes ($60{,}000$ frames) of the main camera footage taken from a game of the German Bundesliga. As the stadium is much bigger than the ones of SoccerNet, we set the default focal point position $\cameraFocalPoint_{default} = (0, 90, -18)$, and we use a commercial tool for keypoints and markings detections~\cite{EVS2022Xeebra}. To derive the tripod position, we run our system without the tripod constraint on the first $5{,}000$ frames, and perform the optimization procedure as described in \Cref{subsec:tracking:results:sn-gamestate-evaluation} to derive $\cameraTripodPoint$ and $\delta$.  This sequence is not annotated;  hence, only the overlay of the soccer field projection provides a qualitative assessment. As displayed in \Cref{fig:tracking:long-sequence}, the projection of the soccer field overlays almost perfectly on top of the actual field markings. \ourMethodName maintains the same quality of overlay for the complete video; this illustrative video is given in the supplementary material. Moreover, the reinitialization step is only performed 60 times, and lasts for 4 frames on average. We explain this improvement by the quality of the commercial keypoints and markings detection.  

\section{Conclusions\label{sec:tracking:Conclusions}}
In this paper, we have presented \ourMethodName, a new tracking system specially designed for wide-angle broadcast cameras. Through diverse qualitative and quantitative evaluation, we show that our system is both accurate and robust, outperforming available solutions in both aspects. Our results also corroborate the suitability of our broadcast camera lens and tripod models, motivating further exploration and refinement. Future works entail the dynamic incorporation of the tripod constraint, even if \ourMethodName obtains convincing results with a roughly appropriate focal point. Another exciting piece of research lies in the vertigo effect noticed with the high-frequency variation of the focal length, as we believe modeling this effect might lead to even better focal point position. 

\paragraph*{Acknowledgments.} This work was supported by the Service Public de Wallonie (SPW) Recherche, Belgium, under Grant $\text{N}^{\text{o}}$8573.

\end{document}